\documentclass[conference]{IEEEtran}
\IEEEoverridecommandlockouts

\usepackage{cite}
\usepackage{amsmath,amssymb,amsfonts,url}
\usepackage{algorithmic}
\usepackage{graphicx}
\usepackage{textcomp}
\usepackage{graphicx,multirow}
\usepackage{enumitem}
\usepackage{comment}
\usepackage{booktabs}
\usepackage{color, colortbl}
\usepackage[usenames,dvipsnames]{xcolor}
\usepackage{makecell}
\usepackage[normalem]{ulem}
\usepackage{adjustbox}
\usepackage{amssymb}
\usepackage{pifont}

\def\BibTeX{{\rm B\kern-.05em{\sc i\kern-.025em b}\kern-.08em
    T\kern-.1667em\lower.7ex\hbox{E}\kern-.125emX}}

\newcommand{\cmark}{\ding{51}}%
\newcommand{\xmark}{\ding{55}}%

\definecolor{shadecolor}{RGB}{0,150,0}

\newcommand{\COMMENT}[1]{}
\begin{document}

\title{Splitformer: An improved early-exit architecture for automatic speech recognition on edge devices\\
}
%


%
%
%



\author{\IEEEauthorblockN{1\textsuperscript{st} Maxence Lasbordes}
\IEEEauthorblockA{\textit{Université Paris-Dauphine, Université PSL,} \\
\textit{Télécom SudParis, Institut Polytechnique de Paris}\\
Paris, France \\
maxence.lasbordes@dauphine.eu}
\and
\IEEEauthorblockN{2\textsuperscript{nd} Daniele Falavigna}
\IEEEauthorblockA{\textit{Center for Augmented Intelligence} \\
\textit{Fondazione Bruno Kessler}\\
Trento, Italy \\
falavi@fbk.eu}
\and
\IEEEauthorblockN{3\textsuperscript{rd} Alessio Brutti}
\IEEEauthorblockA{\textit{Center for Augmented Intelligence} \\
\textit{Fondazione Bruno Kessler}\\
Trento, Italy \\
brutti@fbk.eu}\thanks{ This work was partially funded by the PNRR project CN - HPC (Spoke 9) under the NRRP MUR program funded by the NextGenerationEU. }
}

%

\linespread{0.93}
\maketitle

\begin{abstract} 

The ability to dynamically adjust the computational load of neural models during inference in a resource aware manner is crucial for on-device processing scenarios, characterised by limited and time-varying computational resources. Early-exit architectures represent an elegant and effective solution, since they can  process the input with a subset of their layers, exiting at intermediate  branches (the upmost layers are hence removed from the model). 

From a different perspective, for  automatic speech recognition applications there are memory-efficient neural architectures  that apply  variable frame rate analysis, through downsampling/upsampling operations in the middle layers, reducing the overall number of operations  and improving significantly  the performance on well established benchmarks. One example is the Zipformer. However, these architectures lack the modularity necessary to inject early-exit branches.

With the aim of improving the performance in early-exit models, we propose introducing parallel layers in the architecture   that process downsampled versions of their inputs. 
We  show that in this way the speech recognition performance on standard benchmarks significantly improve, at the cost of a
small increase in the overall number of model parameters but without affecting the inference time.

\end{abstract}
\begin{IEEEkeywords}
conformer, zipformer, early-exit, dynamic models, ASR
\end{IEEEkeywords}
\section{Introduction}
\label{sec:intro}

Although the use of large speech foundation models (SFMs) is very widespread nowadays for automatic speech recognition (ASR) applications, their utilization on edge devices, where memory and computation resources are very limited, is still prevented. Resource-constrained applications require to use models that have much less parameters than those of SFMs and that can dynamically change their trade-off between computation and performance. To this end, we investigated in the past  the use of early-exit architectures applied to large-vocabulary ASR~\cite{wright_icassp_2024}.

``Early-exit'' (EE) architectures  introduce intermediate exit branches~\cite{branchynet2017, phuong2019} that allow to process the input by only a subset of layers exiting before reaching the top layer, thus saving computations. Early-exit architectures leverage the observation that, for simpler inputs, the lowest layers of the model may have already learned sufficient parameters for accurate predictions and have been succesfully applied to ASR~\cite{zaiem2023,xin2022}.

\begin{figure}[h]
  \centering
  \includegraphics[width=1\linewidth]{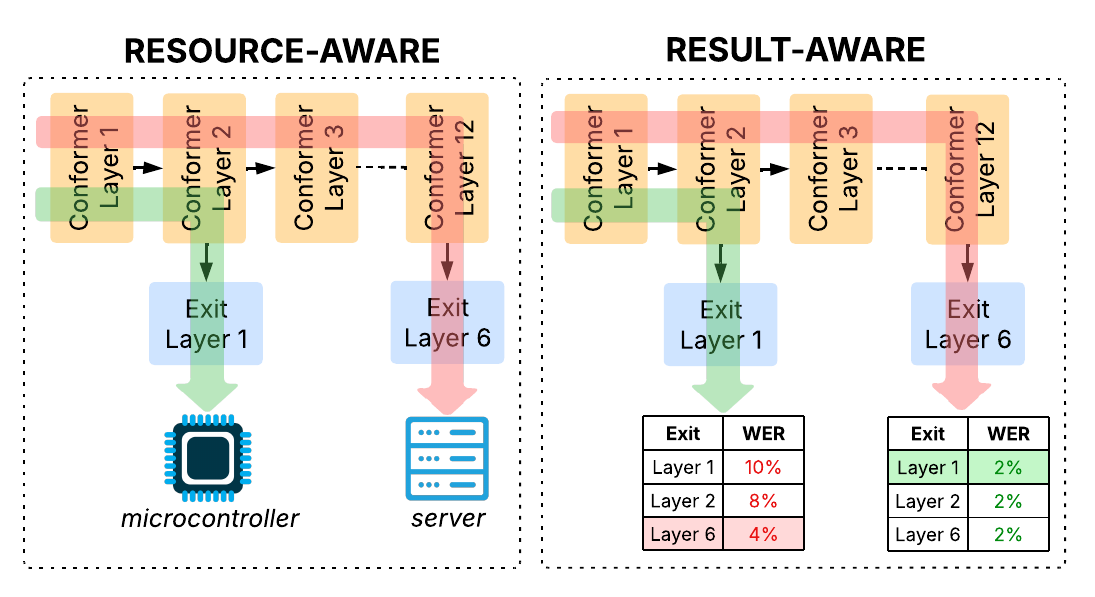}
\caption{\textit{Resource-aware} and \textit{result-aware} uses of early-exits. Left: the device is allowed to dynamically process only two layers, whereas the server can handle the whole model. Right: the first input needs to be processed by by all of the encoder layers; in the second case, the best transcription is produced after only two encoder layers.}
\label{fig:awareness}
\end{figure}

Figure~\ref{fig:awareness} shows an example of an early-exiting network, where specific decoders process exits of intermediate  layers. Note that this architecture is suited to be  implemented either over a distributed environment, consisting of models with a device specific number of layers ("resource aware"), or using  a single model that selects the best exit according to a given metric ("result aware"). In this work, we address only the resource-aware case, referring  the reader to our previous work for details on the usage of early-exit metrics~\cite{wright_icassp_2024}.

Recently the {\em Zipformer}, a faster and more memory-efficient encoder architecture, has been proposed for ASR~\cite{yao2024zipformer}. This architecture overcomes all previous confomer based models~\cite{gulati2020conformer, peng2022branchformer, kim2022squeezeformer} both in terms of performance and computational load (meant  as both  memory occupancy and inference time). The zipformer implements, in addition to various innovative features, a stack of layers that downsample the input sequence with different lower frame rates. The intuition behind this approach is that processing with a few additional layers, applying different sampling rates, allows for broader acoustic contexts to be considered. Unfortunately, these architectures, introducing different frame rates through the backbone, do not marry well with early exit branches.

Inspired by the Zipformer, we have modified our previous early-exit architecture~\cite{wright_icassp_2024}  by exploiting both input layer  downsampling and parallel connections between layers. Experimenting with different variants of this approach, we observed that the best model includes two parallel  downsampling layers, in both the first and last encoder outputs.

We show that in this way the word error rate (WER) on standard ASR benchmarks can be significantly reduced, at the cost of a small increase in the overall number of model parameters. We have also measured small gains in the computation times in the lowest exits of the proposed architecture. 
In summary, the contribution of this work is as follows: {\em i)} the development of a new EE architecture inspired by zipformer; {\em ii)} an experimental analysis on well established ASR benchmarks showing the effectiveness of the proposed architecture; {\em iii)} an analysis of performance vs. model complexity, expressed both in terms of computation times and number of floating point operations (FLOPS), depending on each exit.

\section{Related Work}
\label{sec:related}
Early-exiting methods were first introduced for computer vision in BranchyNet \cite{branchynet2017} by adding two branches to AlexNet \cite{alexnet2017}. 
The authors optimised the joint loss of the exits and defined a confidence measure, based on the entropy of the output class distribution, to decide the exit level. 
%
%
To accelerate inference time in ASR,~\cite{hubert2021} proposes to use confidence measures, given by  CTC decoders, or entropies computed from the logits of early-exits  of a pre-trained audio encoder (i.e. HUBERT). 
A deep analysis of "overthinking" of ASR encoders has been carried out in~\cite{berrebbi2022avoid}, where theoretical lower limits of inference speed vs. ASR performance have been derived for different early exit strategies. Similar investigations have been explored in~\cite{xin2022} for streaming using recurrent neural networks. \cite{zaiem2023} have investigated early-exit fine-tuning strategies in the context of a large pre-trained WavLM~\cite{wavlm2022} model, comparing them with approaches based on layer removal and input down-sampling.

In our previous work~\cite{wright_icassp_2024} we investigated the training dynamics of early-exit models, showing that training the model from scratch,
jointly optimizing all exited layers, provides significant performance improvements over both conventional single-exit models and fine-tuned pre-trained models (particularly at the lowest exits).

From a different perspective, two previous works have optimized the conformer architecture, for ASR tasks,  introducing variants both in several points of the basic conformer module and in the pipeline. The {\em Squeezeformer} architecture described in~\cite{kim2022squeezeformer} uses: {\em a)} the U-net~\cite{unet2015}  temporal structure to reduce the computations required by the multi-head attention modules in the conformer pipeline and {\em b)} a simplified basic block, as an alternative to the Macaron structure of the conformer~\cite{gulati2020conformer},  similar to a standard transformer block. This work shows that  the U-net structure, applying downsampling/upsampling operations to the input/output layers of the architecture, allows to reduce the number of operations in the whole architecture maintaining, at the same time, the resolution of the original input features. Similarly, the {\em Zipformer} architecture~\cite{yao2024zipformer} operates in the middle layers at lower frame rates. It also applies a set of  changes to the basic conformer architecture that improves both  velocity and memory efficiency.  Either architectures have been successfully applied to standard ASR benchmarks  showing significant performance gain.

In this work, we inherit the idea of U-net and apply it to an early-exit architecture. To make this feasible we introduce parallel layers in the baseline model, that perform downsampling/upsampling of the input embeddings. This approach significantly improves the overall performance and also allows to slightly reduce the decoding times in the lowest exits.

\COMMENT{
The issue of overthinking in ASR encoders has been analysed in \cite{berrebbi2022avoid}, in which the authors report theoretical lower bounds of speed/quality trade-offs for early-exit strategies. Exit-selection strategies were proposed based on comparisons of output distribution and transcriptions between successive exits.  

All aforementioned studies~\cite{hubertee2022, berrebbi2022avoid, zaiem2023} employ pre-trained models by fine-tuning the Transformer component, as is common for ASR. They are primarily focused on \textit{improving inference efficiency} by selecting the best early-exit according to some criteria. Analogously, in natural language processing, research on early-exit strategies has focused on accelerating the inference of large pre-trained language models such as BERT~\cite{Zhou2020, Liu2020, Xin2020}. 
}

\section{Early-Exit Loss}
\label{sec:arch}

Given an input sequence of acoustic observations $\mathbf{x}$ and a neural model $\Theta$, an ASR system estimates 
the output sequence $\hat{\mathbf y}=\hat{\mathbf y}_1, \ldots, \hat{\mathbf y}_L$ as:
\begin{equation}
\label{eq:asr}
\hat{\mathbf{y}} = \mathrm{arg}\max_{\mathbf{y}} P[\mathbf{y}|\mathbf{x},\Theta],
\end{equation}
where $\mathbf{y} \in \mathcal{Y}^{\ast}$, for some vocabulary $\mathcal{Y}$, such as graphemes, phonemes, or  word-pieces.  
Usually $\Theta$ is  factored into an encoder and a decoder. The encoder extracts an high-dimensional representation ${\mathbf h}_1^T$ of the input ${\mathbf x}$  and the decoder maps this representation into the output sequence $\hat{\mathbf y}$.
Since $L\ll T$ in general, ASR decoders either use; {\em a)} an \textit{alignment function} ($\mathcal{B}:\mathbf{a}_1^T\rightarrow \mathbf{r}$) for sequence training ($\mathbf{r}$ is the reference  labels string), or {\em b)} a \textit{cross attention mechanism} with label-based cross-entropy optimization~\cite{vaswani2017}.
Our goal is to apply  early-exiting to ASR by adding decoders at some intermediate layers of the encoder (see Fig.~\ref{fig:awareness}). Assuming to use $M$ intermediate decoders, giving outputs $\hat{\mathbf y}(1), \ldots, \hat{\mathbf y}(M)$, the overall model is trained by optimizing the following joint objective:
\begin{equation}
\label{eq:jointloss}
    \mathcal{L}_{EE}(\hat{\mathbf{y}}(1),\ldots,\hat{\mathbf{y}}(M),\mathbf{r}) = \sum_{m=1}^{M}\mathcal{L}(\hat{\mathbf{y}}(m),\mathbf{r}),
\end{equation}
where $\mathcal{L}(\hat{\mathbf{y}}(m),\mathbf{r}) = -\log P[\mathbf{r}|\mathbf{x},\Theta_m]$, and $\Theta_m$ denotes the subset of parameters of $\Theta$ from the first to  the m-th layer.
\COMMENT{
In this work, we implement early-exiting in four diverse model architectures. Conformer-CTC and Conformer-AED share a Conformer encoder~\cite{conformerlu2019} architecture, with early-exits appended to every other layer of a 12-layer encoder, but features different decoders. In addition we consider two pre-trained backbones based on Wav2Vec-2.0~\cite{baevski2020wav2vec2} and WavLM~\cite{wavlm2022} models.
Their hyperparameters are summarised in Tab.~\ref{tab:hyperparams}.
}
In this work the encoder computes $\mathbf{h}_1^T(m), 1\leq m\leq M$, and the decoders  are linear layers with softmax function. They allow to estimate the probability of  $\mathcal{Y} \cup \{\phi\}$ for each input frame, being $\phi$ a "blank" token  indicating "no label issued".
The intermediate loss function is the  connectionist temporal classification (CTC)~\cite{Graves2006ConnectionistTC}, defined as:
\begin{equation}
\label{eq:ctc}
    \mathcal{L}_{\mathrm{CTC}}(\hat{\mathbf{y}},\mathbf{r}) = -\log \sum_{\mathbf{a}_1^T\in\mathcal{B}^{-1}}\prod_{t=1}^T P[a_t|\mathbf{h}_1^T],
\end{equation}
where $a_t \in \mathcal{Y}\cup \{\phi\}$. To ease the notation, in the equation above we have removed the dependence of all the variables on layer $m$.

\section{Proposed Architecture}



While the Zipformer  incorporates numerous blocks and modules specifically designed to optimize ASR performance in a "single exit" architecture, our study focuses solely on the integration of downsampling and upsampling operations within an EE architecture. To assess the effectiveness of this approach, we conducted a preliminary experiment comparing the performance of a "single exit" audio encoder, based on the U-net architecture, with that of a baseline "single exit" conformer-based encoder.

\begin{figure*}[h]
\centering
  \includegraphics[width=0.9\textwidth]{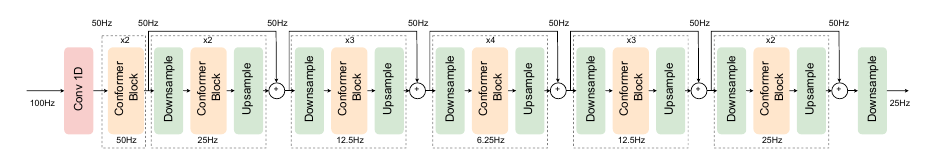}
\caption{The U-net modified architecture of the audio encoder.}
\label{fig:zip1}
\vspace{-0.5cm}
\end{figure*}
Conformer-baseline consists of a 1-D convolutional front end that downsamples the input sequence, represented by 80-dimensional Mel Frequency Cepstral Coefficients (MFCC),  by a factor of two (from 100 Hz to 50 Hz), followed by a positional encoding module that feeds a stack formed by 12  conformer layers. In the "U-net modified" encoder architecture, shown in Figure~\ref{fig:zip1}, the output of the first two conformer layers is sent to  a stack of 5 blocks, each formed by a variable number of conformer layers,  with a residual connection   between them. Similarly to zipformer each block is preceded and followed by downsampling and upsampling operations, respectively, that process the input stream at different frame rates, namely: 50 Hz, 25 Hz, 12.5 Hz and 6.25 Hz. 
In both architectures 
the optimized loss is the CTC loss evaluated from the output of the linear soft-max decoder (note that for the sake of clarity the decoder blocks are not depicted in the figures).

Table~\ref{tab:single_exit} reports the results achieved  with these two architectures on the LibriSpeech benchmark (see Section~\ref{sec:experiments}), where significant improvements yielded by the "U-net modified" encoder can be seen\footnote{
Note that the WERs of the original Zipformer architecture, as reported in~\cite{yao2024zipformer}, are 2.4\% and 5.7\% on test-clean and test-other, respectively. 
The performance discrepancy observed with respect to the two architectures presented in Table ~\ref{tab:single_exit},  can primarily be attributed to the application of data augmentation techniques during the training of the Zipformer — techniques that were not employed when training either the Conformer-baseline or the "U-net modified" architectures.}.
\begin{table}[h]
\vspace{-0.4cm}
\centering

    \caption{\%WERs achieved with the "single-exit" models on the LibriSpeech evaluation data sets.}
    \begin{tabular}{l|l|l}\hline
    {\bf Architecture} & test-clean & test-other\\\hline
    Conformer-baseline &6.1&17.3\\
    U-net modified &4.4&13.3 \\
     \end{tabular}
    \label{tab:single_exit}
\end{table}%
\begin{table*}[h]
    \centering
    \caption{Parameters of the early-exit architectures used in the experiments}
    \setlength{\tabcolsep}{2pt}
    \adjustbox{max width=\linewidth}{
    \begin{tabular}{@{}l|r|r|r|r@{}}
    \toprule
    \textbf{Feature} & \begin{tabular}{r}\textbf{EE-baseline}\end{tabular} & \begin{tabular}{r}\textbf{Splitformer}\textbf{}\end{tabular} & \begin{tabular}{r}\textbf{Wav2Vec2}\textbf{}\end{tabular}& \begin{tabular}{r}\textbf{WavLM (Base+)}\textbf{}\end{tabular} \\
    \midrule
    \# params (M) & 31.5M &  36.7M & 94.0M &94.7M\\
    Encoder & 12-layer Conf. & 14-layer Conf. & 12-layer Transf. & 12-layer Transf.\\
    Attention dim. & 256 &256 &768  & 768\\
    Number heads &8 & 8 & 8 & 8\\
    Feed-forward dim. & 2048 &2048 & 3072 & 3072\\
    Decoder & Linear & Linear & Linear & Linear\\
    Inputs & 80-d MFCC &80-d MFCC & Waveform *  & Waveform * \\
    Loss function & $\mathcal{L}_{\mathrm{CTC}}$ &$\mathcal{L}_{\mathrm{CTC}}$& $\mathcal{L}_{\mathrm{CTC}}$  & $\mathcal{L}_{\mathrm{CTC}}$\\
    Output tokens & BPE & BPE& Grapheme & Grapheme\\
    LM rescoring & \xmark &  \xmark &\xmark& \xmark \\
    Data augmentation & \xmark & \xmark& \cmark& \cmark \\
    \bottomrule
    \end{tabular}}
    \label{tab:hyperparams}
\end{table*}%

\begin{figure}[h]
\centering
  \includegraphics[width=1\linewidth]{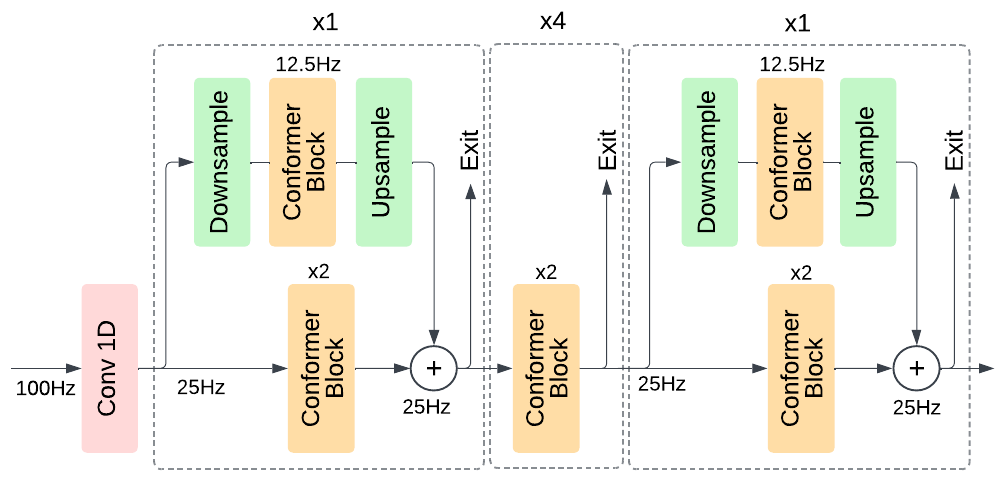}
\caption{The Splitformer architecture of the audio encoder.}
\label{fig:zip_exit}
\end{figure}

Therefore in order to exploit,  similarly to U-net,   variable time resolution processing in an early-exit architecture, we introduce {\em Splitformer}, 
the architecture shown in Figure~\ref{fig:zip_exit}, where: {\em a)}   one exit decoder is inserted every two conformer layers, and {\em b)} the first and last encoder exits are computed by summing the outputs of a standard (two conformer layers) block and of  one parallel layer that processes the input with a downsampling factor equal to $2$.
Despite its simplicity, the experimentation with different configurations, i.e. exploiting a different number of parallel branches and applying different values for the sampling factors, did not produce significant benefits, the architecture of Figure~\ref{fig:zip_exit} being the best compromise between complexity and performance level.


Finally, in the experiments reported below {\em Splitformer} is compared  with {\em EE-baseline}, derived from the Conformer-baseline by simply including one exit decoder every 2 conformer layers. In both of these EE architectures the optimized loss is the one of Equation~\ref{eq:jointloss}. 




\section{Experiments}
\label{sec:experiments}

We  carried out all our experiments  on two well-established public benchmarks for ASR: LibriSpeech~\cite{Libri} and TEDLIUM~\cite{hernandez2018ted} datasets. {\bf LibriSpeech}  contains $\approx$1,000 hours of read-aloud english audiobooks. 
The training set includes 2338 speakers and the evaluation set includes 146 speakers. We refer to \cite{Libri} for further details and baseline performance.  {\bf TEDLIUM-V3} comprises of $\approx$452h of transcribed English speeches from TED video conferences for training and $\approx$6h for evaluation.
\COMMENT{
\begin{table}[h]
    \caption{WER of last exit of all the experimented models.}
    \begin{tabular}{l|l|l|}
    {\bf Model} & text-clean & test-other\\\hline
    Conformer-baseline &5.1&15.1\\
    ZiSplitformer-like (S,M,L) &(6.5,5.4,5.0)&(18.3,16.1,14.7) \\
    Splitformer &5.0&14.4\\
    Wav2Vec2 &4.3&12.2 \\
    WavLM &3.6&8.8 \\
    \end{tabular}
    \label{tab:hyperparams}
\end{table}%
}
Besides comparing our proposed Splitformer with its baseline counterpart (EE-baseline) both trained from scratch, we also experimented with 
two well known pre-trained models, namely: Wav2Vec2~\cite{baevski2020wav2vec2} and WavLM~\cite{wavlm2022}, including early-exits in their architectures and fine-tuning their encoders/decoders layers. Also in this case, the decoders applied in each exit are linear layers followed by soft-max functions.

%
Table~\ref{tab:hyperparams} provides details for all the architectures employed. Differently from the conformer based architectures (i.e. EE-baseline and Splitformer),  both  Wav2Vec2 and WavLM  take as input the raw waveforms and use grapheme base tokenizers with 32 tokens (28 characters + 1 blank token + 2 sentence boundary tokens + 1 unknown token) per their official recipe.
%

\begin{table*}[h]
\centering
\caption{
\%WERs achieved on the on LibriSpeech evaluation sets with the different early-exit models.}
\label{tab:results_libri}
\centerline{
\begin{tabular}{@{}c@{\hskip 3ex}rrrr@{\hskip 3ex}rrrr@{\hskip 3ex}rrrr@{\hskip 3ex}rrrr@{}}
\toprule
\multirow{4}{*}{\textbf{Layer}} &\multicolumn{4}{@{\hskip 3ex}c@{\hskip 3ex}}{\textbf{EE-baseline}}&\multicolumn{4}{@{\hskip 3ex}c@{\hskip 3ex}}{\textbf{Splitformer}}&\multicolumn{4}{@{\hskip 3ex}c@{\hskip 3ex}}{\textbf{Wav2Vec2}}&\multicolumn{4}{@{\hskip 3ex}c@{\hskip 3ex}}{\textbf{WavLM}}\\
\cmidrule(l{1ex}r{1ex}){2-5} \cmidrule(l{1ex}r{1ex}){6-9} \cmidrule(l{1ex}r{1ex}){10-13}  \cmidrule(l{1ex}r{1ex}){14-17} 
& \multicolumn{2}{c}{\texttt{test-clean}} &  \multicolumn{2}{c}{\texttt{test-other}} &  \multicolumn{2}{c}{\texttt{test-clean}} &  \multicolumn{2}{c}{\texttt{test-other}} &  \multicolumn{2}{c}{\texttt{test-clean}} &  \multicolumn{2}{c}{\texttt{test-other}} & \multicolumn{2}{c}{\texttt{test-clean}} &  \multicolumn{2}{c}{\texttt{test-other}} \\
\cmidrule(l{1ex}r{1ex}){2-3} \cmidrule(l{1ex}r{1ex}){4-5} \cmidrule(l{1ex}r{1ex}){6-7} \cmidrule(l{1ex}r{1ex}){8-9} \cmidrule(l{1ex}r{1ex}){10-11} \cmidrule(l{1ex}r{1ex}){12-13} \cmidrule(l{1ex}r{1ex}){14-15} \cmidrule(l{1ex}r{1ex}){16-17} 
  2 && 31.0 && 51.0 &&  28.1  && 48.3 && 33.7 && 56.0 && 28.0 & \ \ \ \ \ \ \ 48.5 &\\
  4 && 11.7 && 27.8 &&  10.8  && 26.4 && 17.4 && 36.7 && 13.9 & 27.3&\\
  6 && 7.1 && 19.8 &&  6.7 && 19.2 && 9.6 && 23.7 && 8.7 & 18.4&\\
  8 && 5.8 && 16.6 &&   5.5  && 16.3 && 5.8 && 15.9 && 4.8 & 12.4&\\
 10 && 5.3 && 15.3 &&  5.1  &&  15.1&& 4.5 && 12.6 && 4.0 &9.5&\\
 12 && 5.1 && 14.8 &&  4.8 && 14.7 && 4.3 && 12.2 && 3.6 & 8.8&\\
\bottomrule 
\end{tabular}
}
\vspace{-0.5cm}
\end{table*}
 Instead, both EE-baseline and Splitformer  use  a byte pair encoding (BPE) based tokenizers~\cite{bpesennrich2016}, with 256 tokens. For their training the learning rate followed the scheduling scheme reported in \cite{vaswani2017}. Specifically, we employed a number of warm-up steps equal to the size of each train dataloader (e.g., $17580$ for the 960h LibriSpeech training set, with batch size equal to $16$). We used the Adam optimiser~\cite{Adam} with $\beta_1=0.9$, $\beta_2=0.98$, and $\epsilon=1e^{-9}$. Dropout, with probability $p=0.1$, was applied before the summation inside each conformer residual block. To limit the number of "pad" tokens in the batches, we removed from each audio dataset training samples having a length greater than 600 characters. We have run $70$ training epochs and for inference we averaged the models of the last 20 epochs. 

To train both Wav2Vec2 and WavLM 
we freezed the features extraction layers of the pre-trained models (we used "Wave2Vec2-base" model) and  fine tuned  optimizing the loss in equation~\ref{eq:jointloss}. The parameters of the Wav2Vec2 model are those of the default configuration  in the huggingface repository\footnote{\url{https://huggingface.co/docs/transformers/model_doc/wav2vec2}}. We trained it using  $50$ epochs.
The code for: EE-baseline, Splitformer  and Wav2Vec2  is available\footnote{\url{https://github.com/augustgw/early-exit-transformer}}\textsuperscript{,}\footnote{\url{https://github.com/augustgw/wav2vec2-ee}},  while the fine-tuning of the WavLM model  follows the related SpeechBrain recipe\footnote{\url{https://github.com/speechbrain/speechbrain}}.

Finally, as mentioned in Section~\ref{sec:intro}, in this work we only  focused on the overall EE model performance, without addressing the task of automatic  exit selection according to some measure of reliability. For this topic we refer the reader to our previous work~\cite{wright_icassp_2024}.

\section{Results}
\label{sec:results}  
Table~\ref{tab:results_libri} shows the results achieved with  all the above mentioned models on the LibriSpeech evaluation data sets.

Notice that the Splitformer delivers superior performance than EE-baseline  in all exits, more consistent in the lowest ones.
Note also that the performance of Splitformer, in the lowest exits, are often better than those achieved with both Wav2Vec2 and WavLM, despite its lower number of parameters. 
In the upmost exits the pre-trained models exhibit significantly better performance, especially on the "test-other"  noisy dataset. However, it has to be considered that, differently from Wav2Vec2/WavLM, Splitformer has not been trained applying data augmentation. Finally, comparing the results of EE-baseline in Table~\ref{tab:results_libri} with those in Table~\ref{tab:single_exit} note  also the superior performance  of the EE loss of equation~\ref{eq:jointloss}, as discussed  in our previous paper~\cite{wright_icassp_2024}.

\begin{table}[th]
\centering
\vspace{-0.4cm}
\caption{
\%WERs obtained with EE-baseline and  Splitformer on the TEDLIUM evaluation sets.}
\label{tab:results_ted}
\centerline{
\begin{tabular}{@{}c@{\hskip 3ex}rrrr@{\hskip 3ex}rrrr@{\hskip 3ex}rr@{\hskip 3ex}rrrr@{}}
\toprule
\multirow{4}{*}{\textbf{Layer}} &\multicolumn{4}{@{\hskip 1ex}c@{\hskip 1ex}}{\textbf{EE-baseline}}&\multicolumn{4}{@{\hskip 1ex}c@{\hskip 1ex}}{\textbf{Splitformer}}\\
\cmidrule(l{1ex}r{1ex}){2-5} \cmidrule(l{1ex}r{1ex}){6-9} 
& \multicolumn{2}{c}{\texttt{dev}} &  \multicolumn{2}{c}{\texttt{test}} &  \multicolumn{2}{c}{\texttt{dev}} &  \multicolumn{2}{c}{\texttt{test}}\\
\cmidrule(l{1ex}r{1ex}){2-3} \cmidrule(l{1ex}r{1ex}){4-5} \cmidrule(l{1ex}r{1ex}){6-7} \cmidrule(l{1ex}r{1ex}){8-9} \cmidrule(l{1ex}r{1ex}){10-11} 
  2 &&45.3  && 45.8 && 37.0   && 37.9  \\
  4 &&20.5 && 21.0 && 18.3   && 18.0  \\
  6 && 13.8&& 13.9 && 13.7  && 13.0  \\
  8 && 11.6 && 11.3 && 11.4  &&  11.4 \\
 10 &&10.8 && 10.6 &&  10.6 &&  10.3 \\
 12 &&10.5 && 10.3 && 10.3  && 9.9\\
\bottomrule 
\end{tabular}
}
\end{table}

When considering only the EE-baseline and Splitformer architectures, the trends observed on the TEDLIUM dataset mirror those reported in Table~\ref{tab:results_libri}, as illustrated in Table~\ref{tab:results_ted}. Also in this case, the performance improvements of Splitformer with respect to the baseline are noticeable, especially in the lowest exits. As a general remark notice that the  improvements in the Splitformer are not only localized in the lowest exit which, due  to the insertion of the parallel downsampling layer, employs 50\% more parameters than the corresponding EE-baseline exit, but also propagate to the upmost exits.

\begin{table*}[t]
\centering

\caption{Total GPU/CPU execution times (in seconds) and number of tera-FLOPS, spent in the encoder, needed to generate the transcriptions in each exit layer. The values have been  computed on the whole evaluation sets of both LibriSpeech and TEDLIUM. The numbers of model parameters before exiting each layer are also given.}
\label{tab:results_time}

\begin{tabular}{c|cc|cc|c||cc|cc|c|}

 & \multicolumn{5}{c||}{\textbf{EE-baseline}} & \multicolumn{5}{c|}{\textbf{Splitformer}} \\
\textbf{Layer} & \multicolumn{2}{c}{{\tt LibriSpeech}} & \multicolumn{2}{c}{{\tt TEDLIUM}} & & \multicolumn{2}{c}{{\tt LibriSpeech}} & \multicolumn{2}{c}{{\tt TEDLIUM}} &\\\hline
& {\bf Time} (sec) & {\bf TFLOPS} & {\bf Time} (sec) & {\bf TFLOPS} & \textbf{\#params} &  {\bf Time} (sec) & {\bf TFLOPS} & {\bf Time} (sec) & {\bf TFLOPS} & \textbf{\#params} \\
 2 & 6112 &10.5 & 3178 &3.3 & 5.4M& 5931 &12.3 &2890  & 3.9&8.0M \\
 4 & 3217 & 20.2& 1801 & 6.4&10.6M & 3168 &22.1 & 1830 &7.0 &13.2M\\
 6 & 2090 &30.0 & 1315 &9.6 & 15.8M& 2339 &31.8 & 1283 & 10.1&18.4M \\
 8 & 1660 &39.7 & 1054 &12.7 &21.1M & 1881 & 41.6& 1144 &13.2 & 23.7M\\
 10 & 1414 & 49.5& 1071 &15.8 &26.3M & 1898 & 51.3& 1080 &16.4 &28.9M\\
 12 & 1343 &59.3 & 1084 &18.9 &31.5M & 1921 &63.5 & 1081 & 20.2&36.7M\\\hline

\end{tabular}
\vspace{-0.5cm}
\end{table*}
\subsection{Computational costs}
Finally, we have carried out some comparisons in terms of the computational costs. Table~\ref{tab:results_time} shows for each exit: the total execution time (including both CPU and GPU times) employed  to generate the automatic  transcripts and the number of tera-FLOPS spent in the corresponding encoder layers. The values of the Table have been  computed on the whole evaluation sets of both LibriSpeech and TEDLIUM.

We observe the following two main results: {\em a)} overall computation times are much higher at lower outputs than at higher ones; {\em b)} the difference in computation times between the two architectures is quite small, the spliformer ones are slightly lower at the lowest exits, while they are a bit higher at the highest exits.
\COMMENT{
\begin{itemize}
    \item overall computation times are much higher at lower outputs than at higher ones;
    \item the difference in computation times between the two architectures is quite small, the spliformer ones are slightly lower at the lowest exits, while they are a bit higher at the highest exits.
   \end{itemize}
}
The result coming from the first point above seems counter-intuitive. In fact one expects that processing time significantly  increases at highest encoder layers, since more operations are needed to traverse all layers. Actually, as expected, the number of FLOPS required by the encoder increases linearly with the number of parameters.
We attribute this behaviour to the pruning operations, applied to the "blank" tokens, in the CTC decoder. The decoder used in our experiments\footnote{\url{https://pytorch.org/audio/main/tutorials/asr_inference_with_cuda_ctc_decoder_tutorial.html}} 
prunes "blank" probabilities higher than 0.95 before expanding hypotheses in the  CTC trellis. Since the number of "blank" tokens is the majority,  their pruning has a strong impact on the overall computation time. Using the EE-baseline model we measured, over the test sets of LibriSpeech: {\em a)}  around 83\% of  tokens generated in the upmost exit are "blank" tokens and {\em b)} those  pruned in the last exit (i.e. above the threshold) are five times those of the first exit.
Therefore,  we speculate that the longer times required to traverse the encoder from bottom to top are counterbalanced by much shorter decoding times at the higher exits.

\section{Conclusions and future works}
In this paper, we have investigated early-exit architectures for ASR by comparing two architectures, one based on a stacked layer encoder, the other employing parallel layers that 
process the input at half of the frame rate. We have proven that the introduction of the parallel layers (the Splitformer architecture)  improves the ASR performance in comparison with: {\em a)} the EE-baseline model and {\em b)} the lowest exits of two foundational pre-trained models.
We have also shown that the computational cost of  Splitformer is comparable with that of EE-baseline.
In future work we will evaluate the effectiveness of the proposed approach on EE models with much fewer parameters (e.g. with lower attention dimension, or reducing the size of the feed-forward network, or the number of attention heads) and experiment on spoken language understanding domains.



\begin{small}

\end{small}

\end{document}